\documentclass[conference]{IEEEtran}
\IEEEoverridecommandlockouts

\usepackage{algorithm}
\usepackage{algpseudocode}
\algtext*{EndWhile}
\algtext*{EndIf}
\algtext*{EndFor}
\algtext*{EndProcedure}
\usepackage{amsmath}
\usepackage{amssymb}
\usepackage{amsthm}
\usepackage{array}
\usepackage{bm}
\usepackage[bookmarks=false]{hyperref}
\usepackage[noadjust]{cite}
\usepackage{caption}
\usepackage{color}
\usepackage{comment}
\usepackage{dblfloatfix}
\usepackage[inline]{enumitem}
\usepackage{epsfig}
\usepackage{etoolbox}
\usepackage{fancybox}
\usepackage{float}
\usepackage{fixltx2e}
\usepackage{graphicx}
\usepackage{grffile}
\usepackage{hyperref}
\usepackage{listings}
\usepackage{mathrsfs}
\usepackage{multirow}
\usepackage{multicol}
\usepackage{pifont}
\usepackage{placeins}
\usepackage{rotating}
\usepackage{setspace}
\usepackage{soul}
\usepackage[subrefformat=parens,farskip=0pt,justification=centering]{subfig}
\usepackage{threeparttable}
\usepackage{times}
\usepackage{url}
\usepackage{upgreek}
\usepackage{verbatim}
\usepackage{wrapfig}
\usepackage[usenames,dvipsnames]{xcolor}
\usepackage{authblk}
\usepackage[]{siunitx}
\usepackage[super]{nth}
\usepackage{tabularx}
\usepackage{supertabular}
\usepackage{tikz}
\usepackage{bbm}
\usepackage{booktabs}
\usepackage{cleveref}

\Crefformat{figure}{Fig.~#2#1#3}                           
\Crefname{subfigure}{Fig.}{Figs.}
\Crefname{figure}{Fig.}{Figs.}
\Crefformat{table}{TABLE~#2#1#3}                           

\newtheorem{Lem}{Lemma}
\newtheorem{Rem}{Remark}
\newcommand{\eat}[1]{}

\graphicspath{{_fig/}}

\newbool{inccomment}
\setbool{inccomment}{true}
\newcommand{\XX}[1]{\ifbool{inccomment}{{\color{magenta} #1}}{}}

\newcommand{\roma}[1]{\uppercase\expandafter{\romannumeral #1\relax}}

\newtheorem{thm}{Theorem}[section] 

\theoremstyle{plain} 
\newcommand{\thistheoremname}{}
\newtheorem{genericthm}[thm]{\thistheoremname}

\newtheorem*{genericthm*}{\thistheoremname}
\newenvironment{namedthm*}[1]
  {\renewcommand{\thistheoremname}{#1}%
   \begin{genericthm*}}
  {\end{genericthm*}}

\title{
    Lithography Hotspot Detection via Heterogeneous Federated Learning with Local Adaptation
}

\author[ ]{
    Xuezhong~Lin$^{1*}$\thanks{$^*$Xuezhong~Lin and Jingyu~Pan contributes equally to this work.},
    Jingyu~Pan$^{2*}$,
    Jinming~Xu$^1$,
    Yiran~Chen$^2$
    and
    Cheng~Zhuo$^1$
}
\affil[$1$]{Zhejiang~University,~China; $^2$Duke~University,~USA}

\begin{document}

\maketitle

\begin{abstract}
As technology scaling is approaching the physical limit, lithography hotspot detection has become an essential task in design for manufacturability.
While the deployment of pattern matching or machine learning in hotspot detection can help save significant simulation time, such methods typically demand for non-trivial quality data to build the model, which most design houses are short of. Moreover, the design houses are also unwilling to directly share such data with the other houses to build a unified model, which can be ineffective for the design house with unique design patterns due to data insufficiency. On the other hand, with data homogeneity in each design house, the locally trained models can be easily over-fitted, losing generalization ability and robustness.
In this paper, we propose a heterogeneous federated learning framework for lithography hotspot detection that can address the aforementioned issues. On one hand, the framework can build a more robust centralized global sub-model through heterogeneous knowledge sharing while keeping local data private. On the other hand, the global sub-model can be combined with a local sub-model to better adapt to local data heterogeneity. The experimental results show that the proposed framework can overcome the challenge of non-independent and identically distributed (non-IID) data and heterogeneous communication to achieve very high performance in comparison to other state-of-the-art methods while guaranteeing a good convergence rate in various scenarios. 

\end{abstract}


\section{Introduction}  \label{sec:intro}
As technology scaling is approaching the physical limit, the lithography process is considered as a critical step
to continue the Moore's law~\cite{moore1965cramming}.
Even though the light wavelength for the process is larger than the actual transistor feature size, recent advances in lithography processing, $e.g.$, multi-patterning, optical proximity correction, $etc.$, have made it possible to overcome the sub-wavelength lithography gap~\cite{2015Optical}. On the other hand, due to the complex design rules and process control at sub-14nm, even with such lithography advances, circuit designers have to consider lithography-friendliness at design stage as part of design for manufacturability (DFM)~\cite{2012Accurate}. 

Lithography hotspot detection (LHD) is such an essential task of DFM, which is no longer optional for modern sub-14nm VLSI designs. Lithography hotspot is a mask layout location that is susceptible to having fatal pinching or bridging owing to the poor printability of certain layout patterns. To avoid such unprintable patterns or layout regions, it is commonly required to conduct full mask lithography simulation to identify such hotspots. While lithography simulation remains as the most accurate method to recognize lithography hotspots, the procedure can be very time-consuming to obtain the full chip characteristics~\cite{2003Hotspot}. To speedup the procedure, pattern matching and machine learning techniques have been recently deployed in LHD to save the simulation time~\cite{2017Layout, 2014A, 2015Machine}. For example, \cite{2014A} built a hotspot library to match and identify the hotspot candidates. Reference \cite{2015Machine} extracted low-dimensional feature vectors from the layout clips and then employed machine learning or even deep learning techniques to predict the hotspots. Obviously, \textit{the performance of all the aforementioned methods heavily depends on the quantity and quality of the underlying hotspot data to build the library or train the model}. Otherwise, these methods may have weak generality especially for unique design patterns or topologies under the advanced technology nodes. 

In practice, each design houses may own a certain amount of hotspot data, which can be homogeneous\footnote[1]{Homogeneous hotspot data refers to the hotspot candidates that share the same feature space due to the similar design patterns or layout topologies.} and possibly insufficient to build a general and robust model/library through \textit{local learning}. On the other hand, the design houses are unwilling to directly share such data with other houses or even the tool developer to build one unified model through \textit{centralized learning} due to privacy concern. 
Recently, advances in federated learning in the deep learning community provide a promising alternative to address the aforementioned dilemma. Unlike centralized learning that needs to collect the data at a centralized server or local training that can only utilize the design house's own data, \textit{federated learning} allows each design house to train the model at local, and then uploads the updated model \textit{instead of data} to a centralized server, which aggregates and re-distributes the updated global model back to each design house~\cite{mcmahan2017communication}.

While federated learning naturally protects layout data privacy without direct access to local data, \textit{its performance (or even convergence) actually can be very problematic when data are heterogeneous (or so-called non-Independent and Identically Distributed, $i.e.$, non-IID)}. However, such heterogeneity is very common for lithography hotspot data, as each design house may have a very unique design pattern and layout topology, leading to lithography hotspot pattern heterogeneity. To overcome the challenge of heterogeneity in federated learning, the deep learning community recently introduced many variants of federated learning~\cite{2020A, smith2018federated, 2017Model, 2018Federated}. For example, federated transfer learning~\cite{2020A} ingested the knowledge from the source domain and reused the model in the target domain. In~\cite{smith2018federated}, the concept of federated multi-task learning is proposed to allow the model to learn the shared and unique features of different tasks. To provide more local model adaptability, \cite{2017Model} used meta-learning to fine-tune the global model to generate different local models for different tasks. \cite{2020Think} further separated the global and local representations of the model through alternating model updates, which may 
get trapped at a sub-optimal solution when the global representation is much larger than the local one. A recent work \cite{2018Federated} presented a framework called FedProx that added a proximal term to the objective to help handle the statistical heterogeneity. Note that LHD is different from the common deep learning applications: LHD is featured with limited design houses (several to tens) each of which usually has a reasonable amount of data (thousands to tens of thousands layout clips). The prior federated learning variants~\cite{2020A, smith2018federated, 2017Model, 2020Think, 2018Federated} are not designed for LHD and hence can be inefficient without such domain knowledge. For example, meta learning appears to loosely ensure the model consistency among the local nodes and hence \textit{fails to learn the shared knowledge for LHD} when the number of local nodes is small, while FedProx strictly enforces the model consistency, yielding \textit{limited local model adaptivity to support local hotspot data heterogeneity}. Thus, it is highly desired to have an LHD framework to properly balance local data heterogeneity and global model robustness. 

To address the aforementioned issues in centralized learning, local learning, and federated learning, in this work, we propose an \textbf{accurate and efficient LHD framework using heterogeneous federated learning with local adaptation}. The major contributions are summarized as follows:
\begin{itemize}
    \item The proposed framework accounts for the domain knowledge of LHD to design {a heterogeneous federated learning framework for hotspot detection}. A local adaptation scheme is employed to make the framework automatically balanced between local data heterogeneity and global model robustness. 
    \item While many prior works empirically decide the low-dimensional representation of the layout clips, we propose {an efficient feature selection method} to automatically select the most critical features and remove unnecessary redundancy to build a more compact and accurate feature representation.
    \item A {heterogeneous federated learning with local adaptation} (HFL-LA) algorithm is presented to handle {data heterogeneity} with a global sub-model to learn shared knowledge and local sub-models to adapt to local data features. A synchronization scheme is also presented to support {communication heterogeneity}.
    \item  We perform a detailed {theoretical analysis} to provide the convergence guarantee for our proposed HFL-LA algorithm and establish the relationship between design parameters and convergence performance.
\end{itemize}%
Experimental results show that our proposed framework outperforms the other local learning, centralized learning, and federated learning methods for various metrics and settings on both open-source and industrial datasets. Compared with the federated learning and its variants~\cite{mcmahan2017communication, 2018Federated}, the proposed framework can achieve 7-11\% accuracy improvement with one order of magnitude smaller false positive rate. Moreover, our framework can maintain a consistent performance when the number of clients increases and/or the size of the dataset reduces, while the performance of local learning quickly degrades in such scenarios. Finally, with the guidance from the theoretical analysis, the proposed framework can achieve a faster convergence even with heterogeneous communication between the clients and central server, while the other methods take 5$\times$ iterations to converge.

\section{Background}

\subsection{Feature Tensor Extraction}
\begin{figure}[tb!]
    \centering  
    \subfloat[]{
        \label{Fig.sub.1}
        \includegraphics[width=0.16\textwidth]{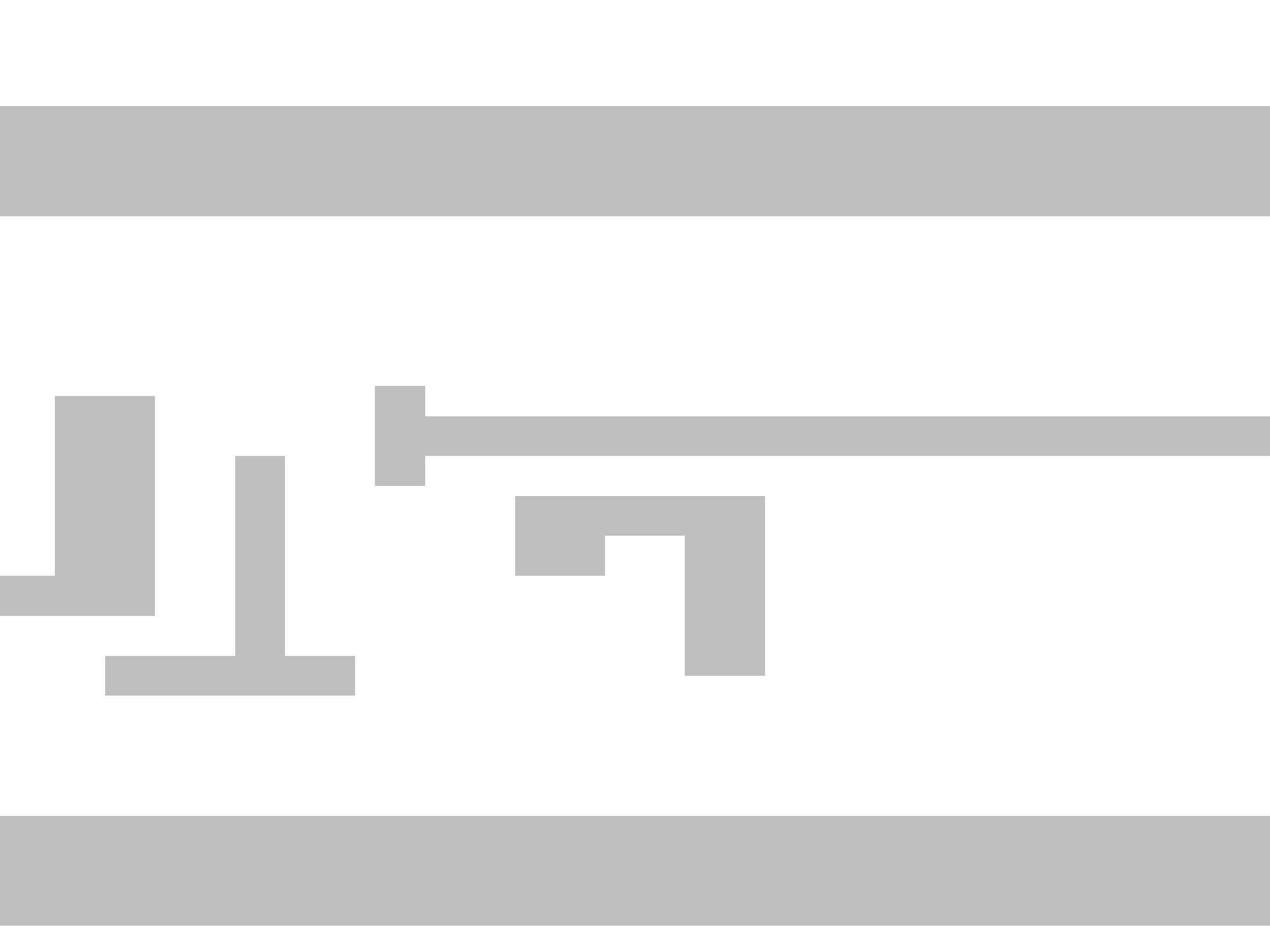}}\subfloat[]{
        \label{Fig.sub.2}
        \includegraphics[width=0.16\textwidth]{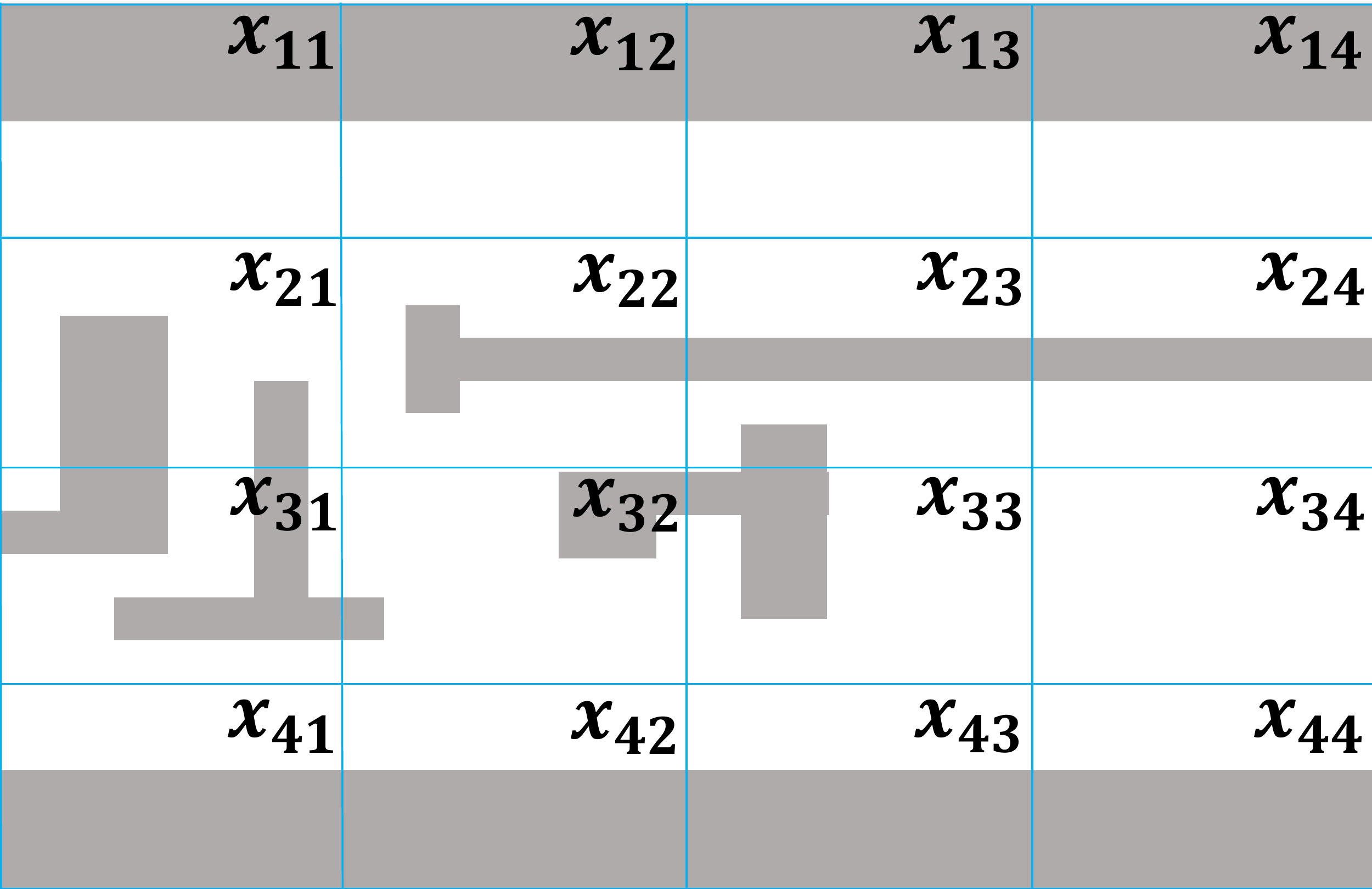}}\subfloat[]{
        \label{Fig.sub.3}
        \includegraphics[width=0.16\textwidth]{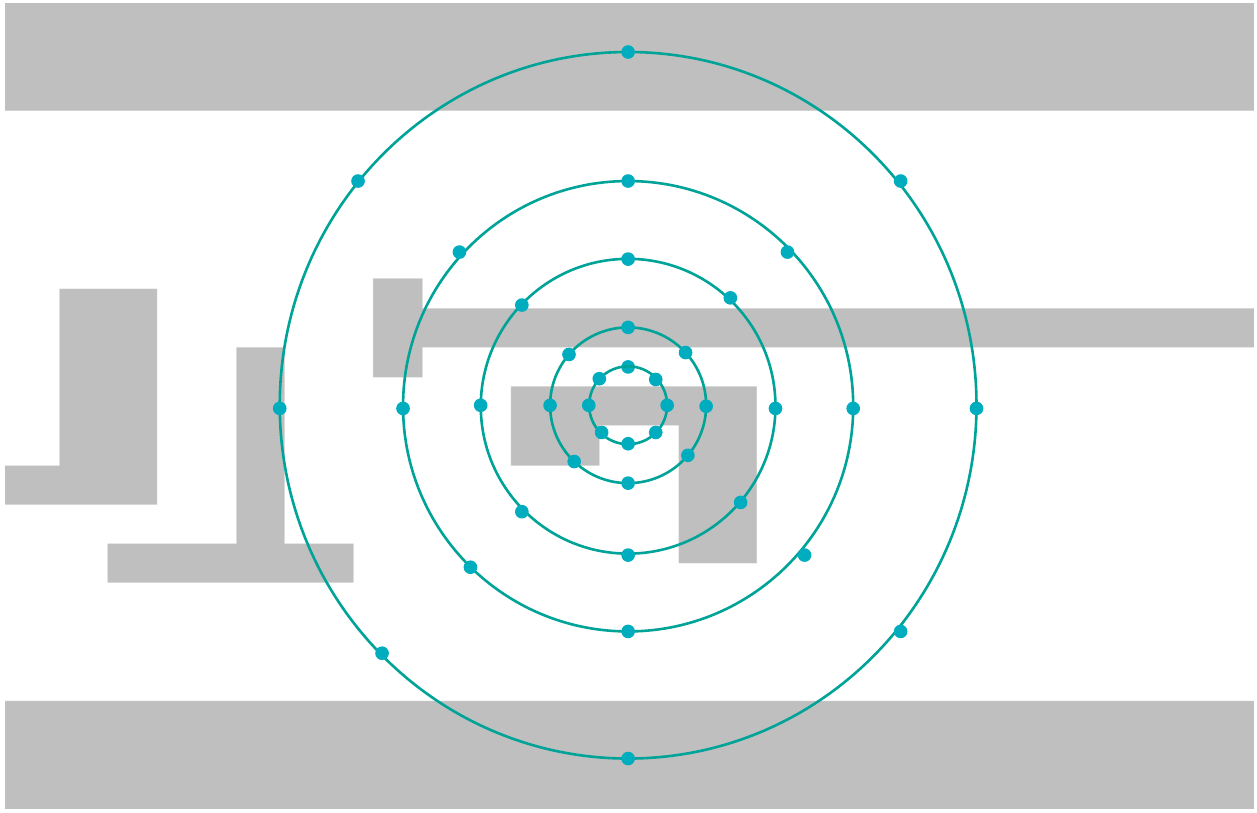}}
    \caption{(a) An example of a layout clip; (b) Local density extraction; (c) Concentric circle sampling.}
    \label{fig_layout_example}
\end{figure}

Feature tensor extraction is commonly used to reduce the complexity of high dimensional data.
For LHD, the original data is hotspot and non-hotspot layout clips composed of polygonal patterns.
Fig.~\ref{fig_layout_example}(a) shows an example of a layout clip.
If unprocessed layout clips are used as features in machine learning, the computational overhead would be huge.
To address this issue, local density extraction and concentric circle sampling have been widely exploited in previous hotspot detection and optical proximity correction works~\cite{matsunawa2015optical, 2017Layout}. 
Fig.~\ref{fig_layout_example}(b) shows an example of local density extraction that converts a layout clip to a vector.
And Fig.~\ref{fig_layout_example}(c) shows an example of concentric circle sampling which samples from the layout clip in a concentric circling manner.
These feature extraction methods exploit prior knowledge of lithographic layout patterns, and hence can help reduce the layout representation complexity in LHD.
However, \textit{as the spatial information surrounding the polygonal patterns within the layout clip are ignored, such methods may suffer from accuracy issues}~\cite{2017Layout}.

Another possible feature extraction is based on the spectral domain~\cite{2017Layout, yang2018layout}, which can include more spatial information. For example, \cite{2017Layout, yang2018layout} use discrete cosine transform (DCT) to convert the layout spatial information into the spectral domain, where the coefficients after the transform are considered as the feature representation of the clip.
Since such feature tensor representation is still large in size and may cause non-trivial computational overhead, \cite{yang2018layout} proposes to ignore the high frequency components, which are supposed to be sparse and have limited useful information. However, such an assumption is not necessarily true for the advanced technologies, which can have subtle and abrupt changes in the shape. In other words, \textit{the ignorance may neglect critical feature components and hence cause accuracy loss}. 


\subsection{Federated Learning}
Federated learning allows local clients to collaboratively learn a shared model while keeping all the training data at local~\cite{mcmahan2017communication}.
Consider a set of $N$ clients connected to a central server, where each client can only access its own local data and has a local objective function $F_k: \mathbb{R}^d \rightarrow \mathbb{R}, k=1,...,N$.
Federated learning can be then formulated as
\begin{equation}
    \min_{w} f(w) = \frac{1}{N} \sum_{k=1}^N F_k(w),
\end{equation}
where $w$ is the model parameter, and $f$ denotes the global objective function.
FedAvg~\cite{mcmahan2017communication} is a popular federated learning method to solve the above problem.
In FedAvg, the clients send updates of locally trained models to the central server in each round, and the server then averages the collected updates and distributes the aggregated update back to all the clients. FedAvg works well with independent and identically distributed (IID) datasets but may suffer from significant performance degradation when it is applied to non-IID datasets.



\section{Proposed Framework} 
\subsection{Overview}
\begin{figure}[tb!]
    \centering
    \subfloat[Procedure for local learning]      { \includegraphics[width=0.92\linewidth]{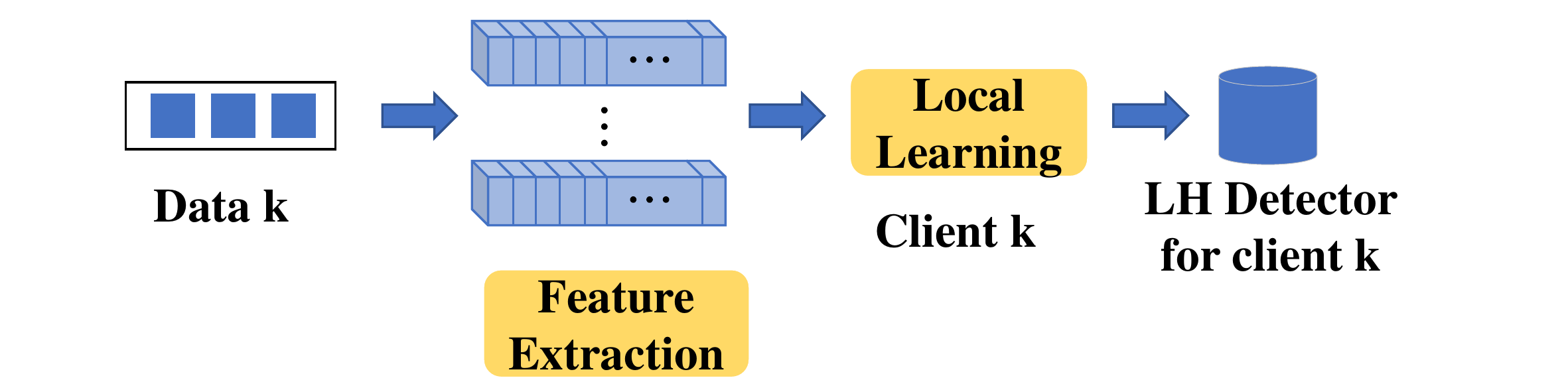} } \\
    \subfloat[Procedure for centralized learning]{ \includegraphics[width=0.45\textwidth]{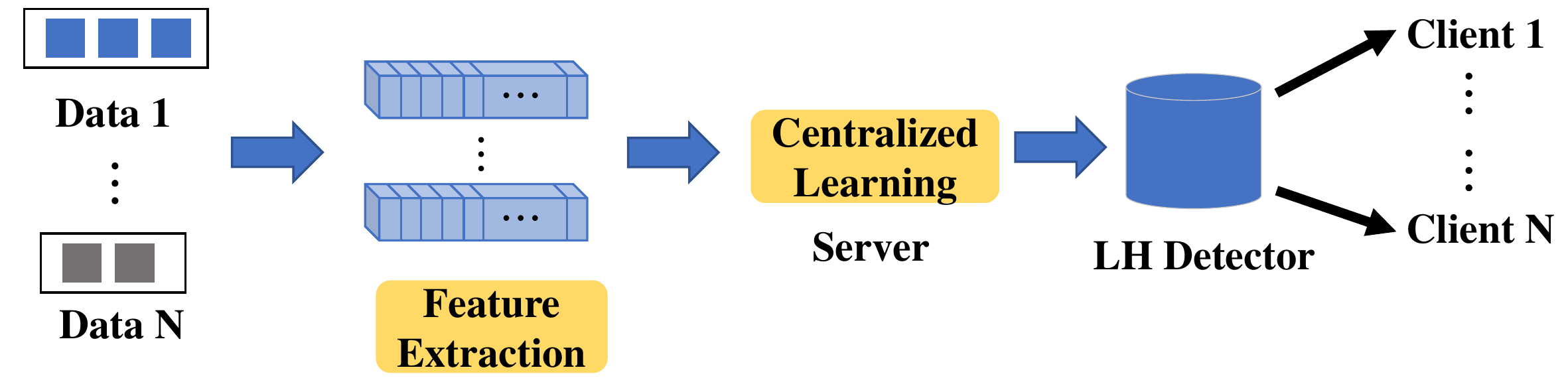} }
    \caption{Two commonly used procedures for LHD.}
    \label{fig:data_learning}
\end{figure}

\Cref{fig:data_learning} demonstrates two commonly used procedures for LHD, $i.e.$, local learning in \Cref{fig:data_learning}(a) and centralized learning in \Cref{fig:data_learning}(b).
Both procedures contain two key steps, feature tensor extraction and learning.
We adopt these two procedures as our baseline models for LHD.
\Cref{tab:nomenclature} defines the symbols that will be used in the rest of the paper.

\begin{table}[b!]
  \centering
  \caption{Symbols used in the proposed framework.}
  \label{tab:nomenclature}
  \resizebox{0.80\linewidth}{!}
  {
      \begin{tabular}{c|l}
     	\toprule
     	Symbol & Definition \\ \midrule
        $w$ & The set of weights of a CNN model \\
         $w_{g}$ & Global weights of the model \\
         $w_{l}^{k}$ & Local weights of the $k_{th}$ client model \\
        $N$ & Total number of clients \\
        $n_{k}$ & The data size of client $k$\\ \bottomrule
      \end{tabular}
  }
\end{table}


The performance of LHD can be evaluated by the true positive rate (TPR), the false positive rate (FPR), and the overall accuracy, which can be defined as follows.

 \begin{namedthm*}{Definition 1} [{\bf True Positive Rate}]
 The ratio between the number of correctly identified layout hotspots and the total number of hotspots.
 \end{namedthm*}

\begin{namedthm*}{Definition 2} [{\bf False Positive Rate}]
 The ratio between the number of wrongly identified layout hotspots (false alarms) and the total number of non-hotspots.
\end{namedthm*}

\begin{namedthm*}{Definition 3} [{\bf Accuracy}]
The ratio between the number of correctly classified clips and the total number of clips.
\end{namedthm*}

With the definitions above, we propose to formulate the following heterogeneous federated learning based LHD:



\begin{namedthm*}{Problem Formulation 1} [{\bf Heterogeneous Federated Learning Based Lithography Hotspot Detection}]
Given $N$ clients (or design houses) owning unique layout data, the proposed LHD is to aggregate the information from the clients and create a compact local sub-model on each client and a global sub-model shared across the clients. The global and local sub-models form a unique hotspot detector for each client.
\end{namedthm*}
The proposed heterogeneous federated learning based LHD aims to support the heterogeneity at different levels: data, model, and communication:
\begin{itemize}
    \item \textbf{Data}: The hotspot patterns at each design house (client) can be non-IID.
    \item \textbf{Model}: The optimized detector model includes global and local sub-models, where the local sub-model can be different from client to client through the local adaptation. 
    \item \textbf{Communication}: Unlike the prior federated learning~\cite{mcmahan2017communication}, the framework allows asynchronous updates from the clients while maintaining good convergence.
\end{itemize}
\eat{There are a few challenges to achieve the above features for LHD:
There are threefold critical challenges.
First, how to filter out the most important characteristics of the hotspot data to be used to build a more effective model.
Second, how to avoid sharing data between clients and build a reliable hotspot detection model to ensure data privacy.
The last but not the least, how to overcome the problem of data non-IID between each client, so as to further maximize the performance of the hotspot detection model.}

\begin{figure*}[tb!]\vspace{-0.2cm}
    \centering
    \includegraphics[width=.9\textwidth]{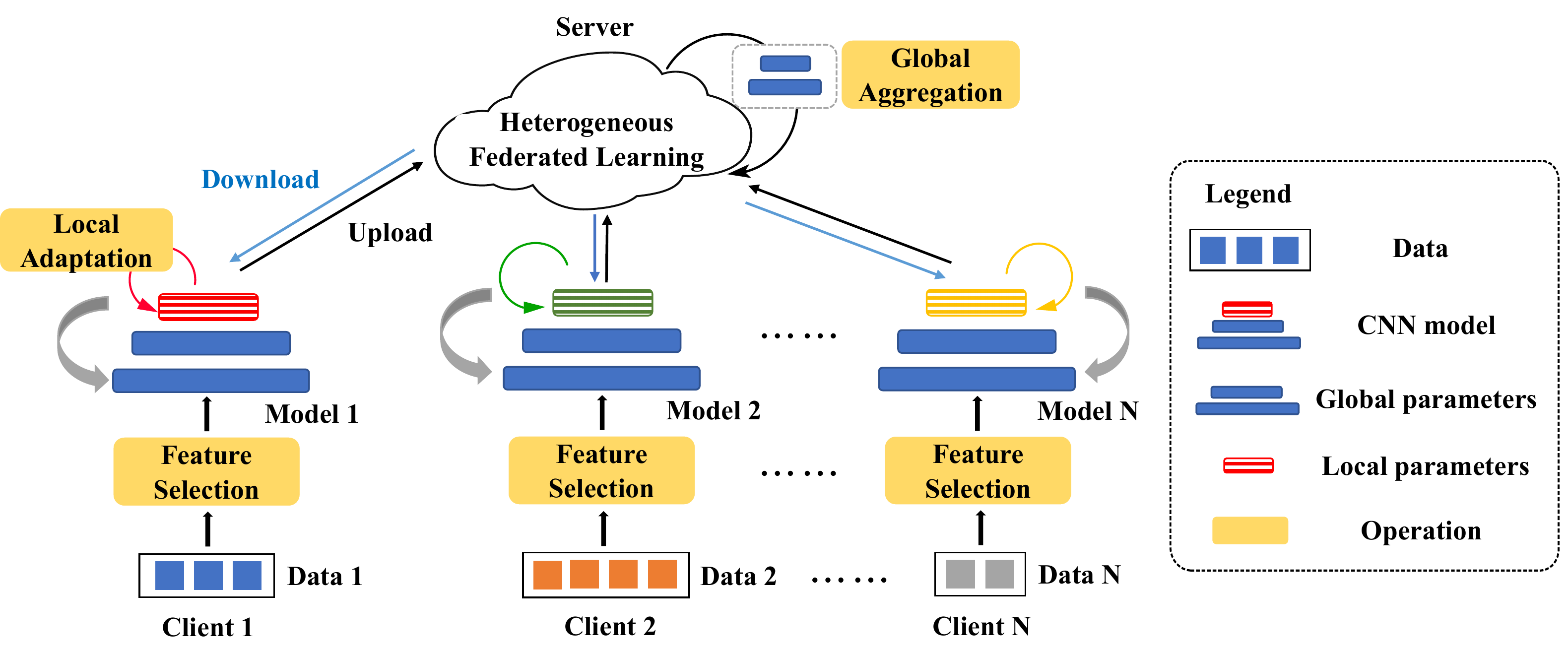}
    \caption{Overview of the proposed LHD framework using heterogeneous federated learning with local adaptation.}
    \label{fig:overview}
\end{figure*}

Figure~\ref{fig:overview} presents an overview of the proposed framework to solve the above LHD problem with the desired features, which includes three key operations: 
\begin{itemize}
    \item \textbf{Feature Selection:} An efficient feature selection method is proposed to automatically find critical features of the layout clip and remove unnecessary redundancy. 
    \item \textbf{Global Aggregation:} Global aggregation only updates the global sub-model shared across the clients with fewer parameters compared to the full model. It does not only reduces the computational cost but also facilitates heterogeneous communication.
    \item \textbf{Local Adaptation:} This operation allows the unique local sub-model at each client to have personalized feature representation of local non-IID layout data. 
\end{itemize}
These operations connect central server and clients together to build a privacy-preserving system, which allows distilled knowledge sharing through federated learning and balance between global model robustness and local feature support. In the following, we will discuss the three operations in details.

\subsection{Feature Selection}
As discussed in Sec.~II-B, while spectral based method can utilize more spatial information, it may easily generate a very large feature vector. To reduce computational cost, the vector is often shortened based on prior knowledge or heuristics~\cite{yang2018layout, 2017Layout}. In this paper, we would like to propose a more automatic feature selection method to find out the most critical components while maintaining the accuracy.

\begin{figure}[b!]
 \centering
 \includegraphics[width=0.45\textwidth]{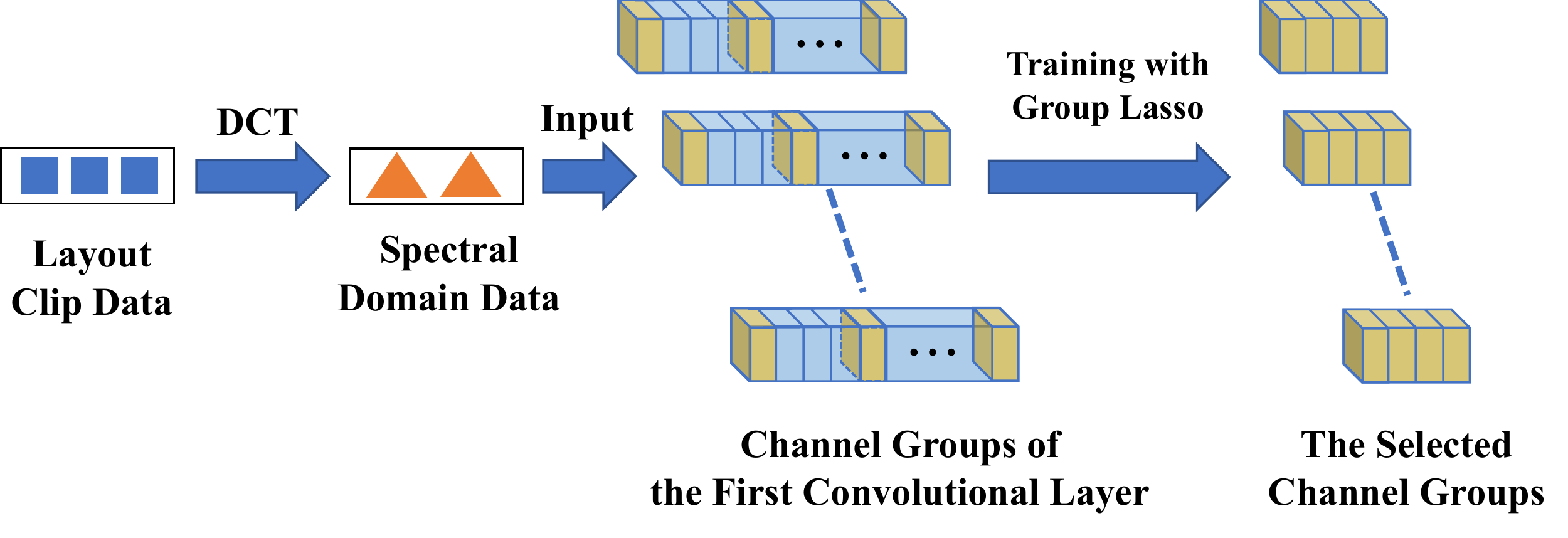}\vspace{-0.2cm}
 \caption{Procedure of the proposed feature selection.}\vspace{-0.2cm}
 \label{fig:FeatureSelection}
\end{figure}

Fig.~\ref{fig:FeatureSelection} shows the proposed feature selection procedure.
The layout clip data is first mapped to the spectral domain with DCT.
Then we use Group Lasso training to remove unwanted redundancy~\cite{yuan2006model}, which is structured regularization to induce grouped sparsity in a deep CNN model.
Generally, the optimization penalized by Group Lasso is
\begin{equation}
    L(w) = L_D(w) + R(w) + \sum_{c=1}^{C} |R_{\ell_2} (w_c)|,
\end{equation}
where $w$ is weights, $L_D(w)$ is cross entropy loss, $R(w)$ is a general regularization term, and $R_{\ell_2}(w_c)$ is structured $\ell_2$ regularization on the $c_{th}$ weight group $w_c$.
In particular, if we make the channels of each filter in the first convolution layer of a deep CNN model a penalized group, the optimization would tend to prune less important channels.
And since each filter channel directly corresponds to a channel in spectral domain features, $i.e.$ a frequency component, this is equivalent to pruning the redundant feature components.
The optimization target with the channel-wise Group Lasso penalty can be defined as
\begin{equation}
    L(w) = L_D(w)
    + \lambda_{R} \left|\left| w \right|\right|_{2}
    + \lambda_{\text{GL}} \sum_{c=1}^{C^{(0)}} \left|\left| w_{:,c,::}^{(0)} \right|\right|_{2},
\end{equation}
where $w$ is the weights of the entire model, $w^{(0)}$ is the weights of the first convolutional layer, $w_{:,c,:,:}^{(0)}$ is the group of the $c_{th}$ channels in all the filters of layer $w^{(0)}$, $\lambda_R$ is the $\ell_2$ regularization strength, and $\lambda_{\text{GL}}$ is the Group Lasso penalty strength.
When the $c_{th}$ feature channel has less impact on reducing $L_{D}(w)$, the penalty term tends to enforce the $\ell_2$-norm of $w_{:,c,::}^{(0)}$ to zero.
Thus, the remaining channels would be the more critical components, leading to a reduction of redundancy in the layout clip feature representation.
Note that we assign the $w^{(0)}$ used for feature selection to the global parameters shown in Fig.~\ref{fig:overview}, the selection result is thus naturally shared by all clients.
Fig.~\ref{fig:CNN} shows an example of the CNN model in our framework.

\subsection{Global Aggregation and Local Adaptation}

Global aggregation and local adaptation are the key operations in the proposed Heterogeneous Federated Learning with Local Adaptation algorithm (HFL-LA). The algorithm HFL-LA is particularly designed for LHD with awareness of its unique domain knowledge: (1) The design patterns in different clients (design houses) may have non-trivial portion in common, which indicates a larger global sub-model for knowledge sharing; (2) The number of clients may be limited, e.g., from several to tens; (3) The local data at each client may be insufficient to support a large local sub-model training. 

As shown in Fig.~\ref{fig:overview}, HFL-LA adopts a flow similar to the conventional federated learning that has a central server to aggregate the information uploaded from the distributed clients. However, unlike the conventional federated learning, the model that each client maintains can be further decomposed into a global sub-model and a local sub-model, where: (1) the global sub-model is downloaded from the server and shared across the clients to fuse the common knowledge for LHD, and (2) the local sub-model is maintained within the client to adapt to the non-IID local data and hence, varies from client to client. 


To derive such a model, we define the following objective function for optimization:
\begin{equation}
    \mathop {\min} \limits_{w_{g},w_{l}}\left\{F\left(w_{g},w_{l} \right) \triangleq\sum_{k=1}^N{p_kF_k\left(w_{g},w_{l}^{k} \right)} \right\} ,
\end{equation}
where $w_{g}$ is the global sub-model parameter shared by all the clients; $w_{l}:=\left[ w_{l}^{1},\cdots ,w_{l}^{N} \right] $ is a matrix whose $k_{th}$ column is the local sub-model parameter for the $k_{th}$ client; $N$ is the number of clients; $p_k\geqslant 0$ and $\sum_{k=1}^N{p_k}=1$ is the contribution ratio of each client; $n_k$ is the data size of client $k$. By default, we can set $p_k=\frac{n_k}{n}$, where $n=\sum_{k=1}^N{n_k}$ is the total number of samples across all the clients. For the local data at client $k$, $F_k\left( \cdot \right) $ is the local (potentially non-convex) loss function, which is defined as
\begin{equation}
    F_k\left( w_{g},w_{l}^{k} \right) = \frac{1}{n_k}\sum_{j=1}^{n_k}{\ell \left( w_{g},w_{l}^{k};x_{k,j} \right)},
\end{equation}
where $x_{k,j}$ is the $j_{th}$ sample of client $k$. As shown in Algorithm~\ref{alg:HFL-LA}, in the $t$ round, the central server broadcasts the latest global sub-model parameter $w_{t,g}$ to all the clients. Then, each client ($e.g.$, $k_{th}$ client) starts with $w_{t}^{k}=w_{t,g}\cup w_{t,l}^{k}$ and conducts $E_{l}\left( \geqslant 1 \right) $ local updates for sub-model parameters
\begin{equation}
\label{eq:interm}
w_{t+\frac{1}{2},l}^{k}=w_{t,l}^{k}-\eta \sum_{i=0}^{E_l-1}{\nabla _lF_k\left( w_{t,g},\hat{w}_{t+i,l}^{k};\xi _{t}^{k} \right)},
\end{equation}
where $\hat{w}_{t+i,l}^{k}$ denote the intermediate variables locally updated by client $k$ in the $t$ round; $\hat{w}_{t,l}^{k}=w_{t,l}^{k}$; $\xi _{t}^{k}$ are the samples uniformly chosen from the local data in the $t$ round of training. After that, the global and local sub-model parameters at client $k$ become $w_{t+\frac{1}{2}}^{k}=w_{t,g}\cup w_{t+\frac{1}{2},l}^{k}$ and are then updated by $E$ steps of inner gradient descent as follows:

\begin{equation}
w_{t+1}^{k}=w_{t+\frac{1}{2}}^{k}-\eta \sum_{i=0}^{E-1}{\nabla F_k\left( \hat{w}_{t+\frac{1}{2}+i}^{k};\xi _{t}^{k} \right)},
\end{equation}
where $\hat{w}_{t+\frac{1}{2}+i}^{k}$ denote the intermediate variables updated by client $k$ in the $t+\frac{1}{2}$ round; $\hat{w}_{t+\frac{1}{2}}^{k}=w_{t+\frac{1}{2}}^{k}$.
Finally, the client sends the global sub-model parameters back to the server, which then aggregates the global sub-model parameters of all the clients, $i.e.$, $w_{t+1,g}^{1},\cdots ,w_{t+1,g}^{N}$, to generate the new global sub-model, $w_{t+1,g}$.

\begin{figure}[b!]
    \centering
    \includegraphics[width=3.4in]{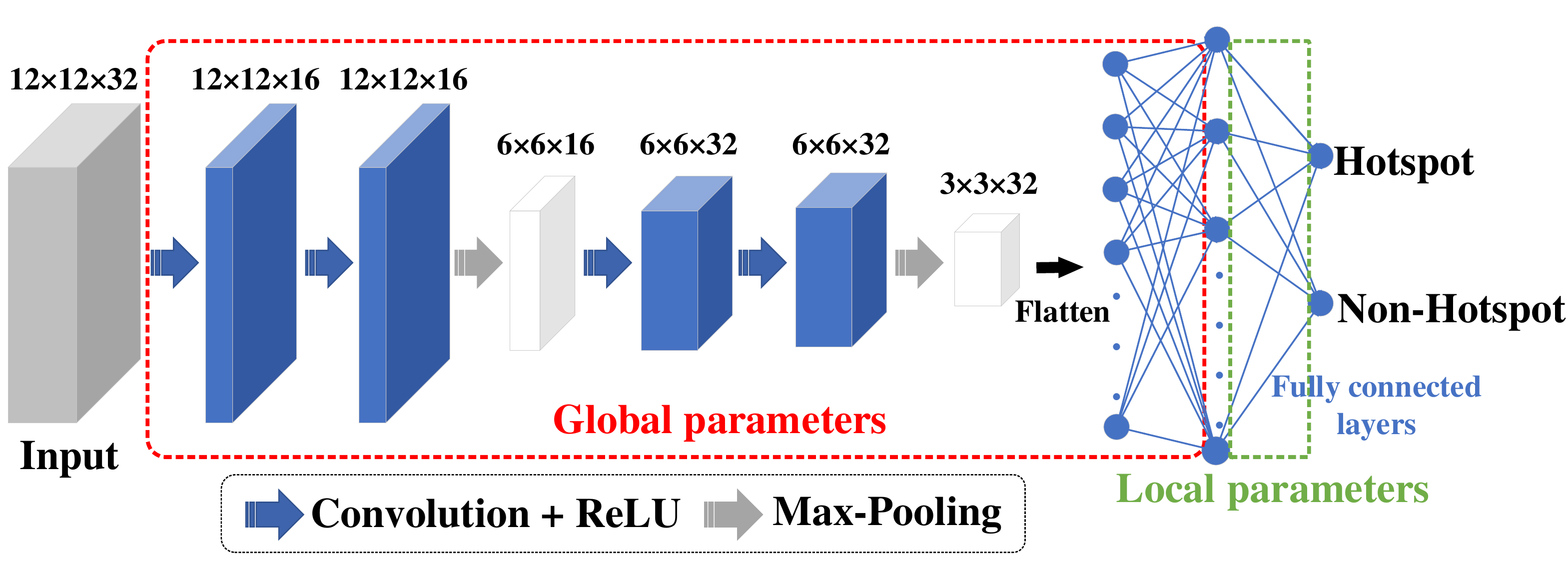}
    \caption{Neural network architecture example at the client.}
    \label{fig:CNN}
\end{figure}

\begin{algorithm}[!bth]
\caption{HFL-LA algorithm for LHD}
\label{alg:HFL-LA}

\textbf{Server}: 
\begin{algorithmic}[1]
\State   Initialize $w _{0,g}$, send $w _{0,g}$ to every client;
    \For {each round $t=0,1,\cdots ,T-1$}
\State      $S_t\gets $ (Randomly select $K$ cilents);
      \For {each client $k\in S_t$ }
\State       $w _{t+1,g}^{k}$ $\gets $ ClientUpdate($k$, $w _{t,g}$);
     \EndFor
\State      $w _{t+1,g}$ $\gets $ $\sum\nolimits_{k=1}^K{\frac{n_k}{n_K}}w _{t+1,g}^{k}$;
\State      Send $w _{t+1,g}$ to every client.
\EndFor
\end{algorithmic}

\textbf{Client}: 
\begin{algorithmic}[1]
\State  ClientUpdate($k$, $w _{g}$):  
\State   $\mathcal{B}\longleftarrow $ (Divide $\mathcal{D}_k$ according to the batch size of $B$);
\For {each local update $i=0,1\cdots ,E_{l}$}  
    \For {batch $\xi ^k\in \mathcal{B}$}
\State        $w_{l}^{k}\gets w _{l}^{k}-\eta \nabla_lF_k(w_{l}^{k};\xi ^k)$;
\EndFor
\EndFor

\For {each global update $i=0,1\cdots ,E$}    
   \For {batch $\xi ^k\in \mathcal{B}$}
\State  $w_{g}^{k}\cup w_{l}^{k}\gets w_{g}\cup w_{l}^{k}-\eta \nabla F_k(w_{g}\cup w_{l}^{k};\xi ^k)$;
\EndFor
\EndFor
\State     return $w _{g}^{k}$ to server.
\end{algorithmic}
\end{algorithm}

Fig.~\ref{fig:CNN} presents the network architecture of each client used in our experiment. The architecture has two convolution stages and two fully connected stages. Each convolution stage has two convolution layers, a Rectified Linear Unit (ReLU) layer, and a max-pooling layer. The second fully connected layer is the output layer of the network in which the outputs correspond to the predicted probabilities of hotspot and non-hotspot. We note that the presented network architecture is just a specific example for the target application and our proposed framework is not limited by specific network architectures. 

\subsection{Communication Heterogeneity}
In addition to data heterogeneity, the proposed framework also supports communication heterogeneity, $i.e.$, the clients can conduct synchronized or asynchronized updates, while still ensuring good convergence. For the synchronized updates, all the clients participate in each round of global aggregation as:
\begin{equation}
w_{t+1,g}=\sum_{k=1}^N{p_kw_{t+1,g}^{k}}.
\end{equation}
Then all the clients need to wait for the slowest client to finish the update. Due to heterogeneity of data, the computational complexity and willingness to participate in a synchronized or asynchronized update may vary from client to client. Thus, it is more realistic to assume that different clients may update at different rates. We can set a threshold $K\left( 1\leqslant K<N \right) $ and let the central server collect the outputs of only the first $K$ responded clients. After collecting $K$ outputs, the server stops waiting for the rest clients, $i.e.$, the $\left( K+1 \right) _{th}$ to $N_{th}$ clients are ignored in this round of global aggregation. Assuming $S_t\left( |S_t|=K \right) $ is the set of the indices of the first $K$ clients in the $t_{th}$ round, the global aggregation can then be rewritten as

\begin{equation}
w_{t+1,g}=\frac{n}{n_K}\sum_{k\in S_t}{p_kw_{t+1,g}^{k}},
\end{equation}
where $n_K$ is the sum of the sample data volume of the first $K$ clients and $\frac{n}{n_K}\sum_{k\in S_t}{p_k}=1$. 

\eat{

\begin{namedthm*}{Partial client participation} [{\bf asynchronous}]
This strategy is much more realistic because it does not require all the clients’ output. We can set a threshold $K\left( 1\leqslant K<N \right) $ and let the central server collect the outputs of the first $K$ responded clients. After collecting $K$ outputs, the server stops waiting for the rest; the $K+1\text{-}\mathrm{th}$ to $N\text{-}\mathrm{th}$ clients are regarded stragglers in this iteration. Let $S_t\left( |S_t|=K \right) $ be the set of the indices of the first $K$ responded clients in the $t\text{-}\mathrm{th}$ iteration. The aggregation step performs:

\begin{equation}
w_{t+1,g}\gets \frac{n}{n_K}\sum_{k\in S_t}{p_kw_{t+1,g}^{k}},
\end{equation}
Where, $n_K$ is the sum of the sample data volume of the first $K$ clients, It can be proved that$\frac{n}{n_K}\sum_{k\in S_t}{p_k}=1$.
\end{namedthm*}

}


\section{Convergence Analysis}
In this section, we study the convergence of the proposed HFL-LA algorithm. Unlike the conventional federated learning, our proposed HFL-LA algorithm for LHD works with fewer clients, smaller data volume, and non-IID datasets, making the convergence analysis more challenging. Before proceeding into the main convergence result, we provide the following widely used assumptions on the local cost functions $\{F_k\}$ and stochastic gradients~\cite{2019Parallel}.


\begin{namedthm*}{Assumption 1}  
\label{assumption:1} 
$F_1,\cdots,F_N$ are all $L\text{-}smooth$, i.e.,  $\forall v, w$, $\left\| \nabla F_k\left( v \right) -\nabla F_k\left( w \right) \right\| \leqslant L\left\| v-w \right\|, \forall k=1,...,N$.
\end{namedthm*}

 \begin{namedthm*}{Assumption 2} Let $\xi_{i}^k$ be uniformly sampled from the $k_{th}$ client's local data. The variance of stochastic gradients in each client is upper bounded, i.e., $\mathbb{E}\left\| \nabla F_k\left( w_{i}^{k};\xi _{i}^{k} \right) -\nabla F_k\left( w_{i}^{k} \right) \right\| ^2\leqslant \sigma ^2$.
\end{namedthm*}

 \begin{namedthm*}{Assumption 3}  
The expected squared norm of stochastic gradients is uniformly bounded by a constant $G^2$, i.e., $\mathbb{E}\left\| \nabla F_k\left( w_i^k;\xi_i ^k \right) \right\| ^2\leqslant G^2$ for all $k=1,\cdots ,N$.
\end{namedthm*}



With the above assumptions, we are ready to present the following main results of the convergence of the proposed algorithm. \textit{The detailed proof can be found in the Appendix.
}

\begin{Lem}[Consensus of global sub-model parameters]\label{lem:bounded_consensus}
Suppose Assumption 3 holds.
Then,
\begin{equation}
\begin{aligned}
\mathbb{E}\left\| \frac{1}{N}\sum_{j=1}^N{w_{t,g}^{j}}-w_{t,g}^{k} \right\| ^2\leqslant \eta ^2\left( E-1 \right) ^2G^2.
\end{aligned}
\end{equation}
\end{Lem}

The above lemma guarantees that the global sub-model parameters of all the clients reach consensus with an error proportional to the learning rate $\eta$ while the following theorem ensures the convergence of the proposed algorithm.

\begin{namedthm*}{Theorem 1}
Suppose Assumption 1-3 hold. Then, $\forall T>1$, we have
\begin{equation}
\begin{aligned}
&\frac{1}{T}\sum_{t=0}^{T-1}{\frac{1}{N}\sum_{k=1}^N{\left[ \left\| \nabla F_k\left( w_{t}^{k} \right) \right\| ^2 \right]}}
\\
&~~~\leqslant \frac{2\left[ \frac{1}{N}\sum_{k=1}^N{F_k\left( w_{0}^{k} \right)}-F^* \right]}{T\eta}
\\
&~~~+\mathcal{O}\left( \eta LG^2 \right) +2\sqrt{N}\left( E-1 \right) G\left( \sigma ^2+G^2 \right). 
\end{aligned}
\end{equation}
\end{namedthm*}
\begin{Rem}
The above theorem shows that, with a constant step-size,  the parameters of all clients converge to the $\eta$-neighborhood of a stationary point with a rate of $\mathcal{O}\left( 1/T \right)$. It should be noted that the second term of the steady-state error is proportional to the square root of $N$, but will vanish when $E=1$. This theorem sheds light on the relationship between design parameters and convergence performance, which helps guide the design of the proposed HFL-LA algorithm.
\end{Rem}


\section{Experimental Results}

\begin{table}[!t]
  \centering
  \caption{Details of the benchmarks, ICCAD and Industry.}
  \label{tab:benchmark}
  \resizebox{\linewidth}{!}{
      \begin{tabular}{c|c|c|c|c|c}
          \toprule
          \multirow{2}{*}{Benchmarks} &
          \multicolumn{2}{c|}{Training Set} &
          \multicolumn{2}{c|}{Testing Set} &
          \multirow{2}{*}{Size/Clip ($\mu m^2$)} \\\cline{2-5}
          & HS\# & non-HS\# & HS\# & non-HS\# & \\
          \midrule
          \texttt{ICCAD} & 1204 & 17096 & 2524 & 13503 & $3.6 \times 3.6$ \\
          \texttt{Industry} & 3629 & 80299 & 942 & 20412 & $1.2 \times 1.2$ \\
          \bottomrule
      \end{tabular}
  }
\end{table}

We implement the proposed framework using the PyTorch library~\cite{paszke2019pytorch}.
We use the following hyperparameters to conduct model training on each client in our experiment:
We train our models with Adam optimizer for $T=50$ rounds with a fixed learning rate $\eta=0.001$ and a batch size of $64$.
And in each round, we conduct local updates for $E_l=500$ iterations, and global updates for $E=1500$ iterations.
To prevent overfitting, we use L2 regularization of $0.00001$.
We adopt two benchmarks (\texttt{ICCAD} and \texttt{Industry}) for training and testing.
We merge all the 28nm patterns in the test cases published in ICCAD 2012 contest~\cite{torres2012iccad} into a unified benchmark denoted by \texttt{ICCAD}.
And \texttt{Industry} is obtained from our industrial partner at 20nm technology node.
Table~\ref{tab:benchmark} summarizes the benchmark details including the training/testing as well as the layout clip size. In the table, columns ``HS\#" and ``non-HS\#" list the total numbers of hotspots and non-hotspots, respectively. Since the original layout clips have different sizes, clips in \texttt{ICCAD} are divided into nine blocks to have a consistent size as \texttt{Industry}. We note that, due to the different technologies and design patterns, the two benchmarks have different feature representations, and \texttt{Industry} has more diverse design patterns ($i.e.$, higher data heterogeneity) than \texttt{ICCAD}.

\eat{
\begin{table}[]
  \centering
  \caption{Training configurations.}
  \label{tab:Configurations}
\begin{tabular}{|c|c|}
\hline
Configurations                      & Value \\ \hline
Optimizer                           & Adam  \\
Learning Rate ($\eta $) & 0.001 \\
Batch Size                          & 64    \\
Feature Tensor Channel              & 32    \\
Training Round ($T$)                  & 50    \\
Local Epoch ($E_l$)                  & 1     \\
Global Epoch ($E$)                    & 3     \\
Step of 1 Epoch                     & 500  \\ \hline
\end{tabular}
\end{table}

}

\begin{figure}[tb!]
    \centering
    \includegraphics[width=1.\linewidth]{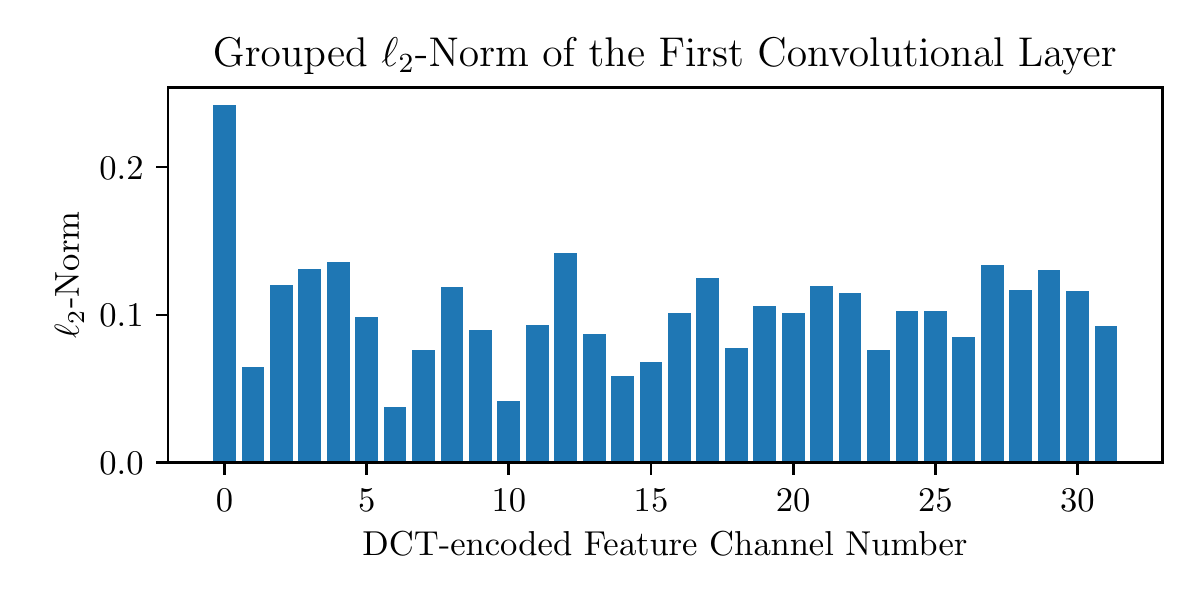}
    \caption{Grouped $\ell_2$-norm of the first convolution layer. The DCT-encoded channel number spans from $0$ to $31$, where the channel $0$ denotes the DC component of the spectral domain data, and the channels $1$-$31$ denote AC components of increasing frequency.}
    \label{fig:ft-importance}
\end{figure}

\begin{figure}[t!]
    \centering
    \includegraphics[width=1.\linewidth]{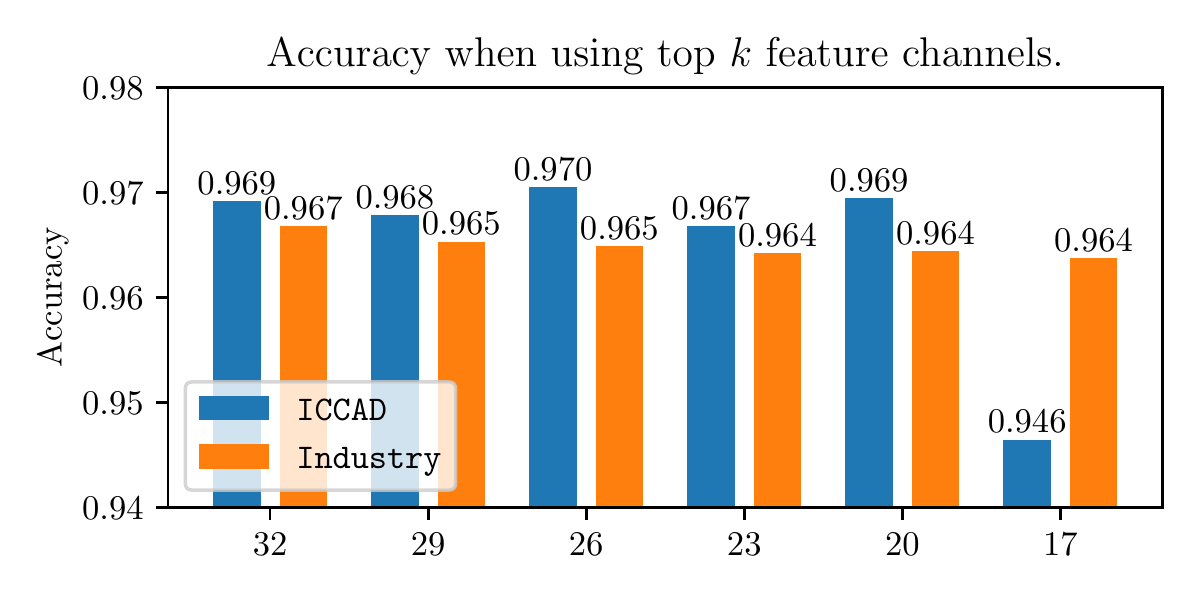}
    \caption{Accuracy of HFL-LA on the validation set using different number of features representing the layout clip.}
    \label{fig:ft_sel}
\end{figure}

\subsection{Feature Selection}

This subsection presents the performance of the proposed feature selection method. As discussed in Sec.~III-B, L2 norm of the channel-wise groups in the first convolutional layer is correlated with the contributions to model performance from the corresponding feature channels, as shown in Fig.~\ref{fig:ft-importance}.
We then sort all the feature channels by their L2 norms and retrain our model from scratch with the selected top-$k$ channels, $i.e.$, $k=26$ in the experiment.
To validate the efficiency of our feature selection method, we test the performance of HFL-LA with different numbers of features representing the layout clips on the validation set and compare the performance.
Fig.~\ref{fig:ft_sel} shows that HFL-LA achieves comparable (even slightly higher) accuracy in the case of $k=26$ features as suggested by the proposed selection method for both benchmarks, which indicates a $18.75\%$ computation reduction for the following learning in comparison to the original 32 features.

\subsection{Heterogeneous Federated Learning with Local Adaptation}


\begin{table}[b!]
    \centering
    \caption{Inference performance (TPR, FPR and accuracy) comparison among HFL-LA, FedAvg, FedProx, local and central learning.}
    \label{tab:SynchronousCommunication}
    \resizebox{\linewidth}{!}{
        \begin{tabular}{c|c|ccc|ccc}
            \toprule
            \multicolumn{1}{c|}{\multirow{2}{*}{{Methods}}} & \multicolumn{1}{c|}{\multirow{2}{*}{{\begin{tabular}[c]{@{}c@{}}Number of \\ clients\end{tabular}}}} & \multicolumn{3}{c|}{{ICCAD}}                             & \multicolumn{3}{c}{{Industry}}     \\ \cline{3-8} 
                \multicolumn{1}{c|}{}                                  & \multicolumn{1}{c|}{}                                                                                       & {TPR} & {FPR} & \multicolumn{1}{c|}{{ACC}} & {TPR} & {FPR} & {ACC} \\ \midrule
                \multirow{3}{*}{{HFL-LA}}                       & {2 clients}                                                                                          & 0.960        & 0.019        & \textbf{0.980}                    & 0.966        & 0.040         & 0.964        \\
                & {4 clients}                                                                                          & 0.967        & 0.021        & \textbf{0.979}                             & 0.975        & 0.049        & \textbf{0.968}         \\
                & {10 clients}                                                                                         & 0.967        & 0.030         & 0.970                              & 0.971        & 0.050         & \textbf{0.965}        \\ \midrule
                \multirow{3}{*}{{FedAvg}}                       & {2 clients}                                                                                          & 0.974        & 0.110         & 0.892                             & 0.814        & 0.010         & 0.869        \\
                & {4 clients}                                                                                          & 0.971        & 0.101        & 0.901                             & 0.883        & 0.016        & 0.914        \\
                & {10 clients}                                                                                         & 0.969        & 0.090         & 0.911                             & 0.881        & 0.016        & 0.913        \\ \midrule
                \multirow{3}{*}{{FedProx}}                      & {2 clients}                                                                                          & 0.977        & 0.134        & 0.868                             & 0.854        & 0.014        & 0.895        \\
                & {4 clients}                                                                                          & 0.973        & 0.121        & 0.880                              & 0.859        & 0.017        & 0.898        \\
                & {10 clients}                                                                                         & 0.958        & 0.113        & 0.888                             & 0.843        & 0.016        & 0.887        \\ \midrule
                \multirow{3}{*}{{Local}}                        & {2 clients}                                                                                          & 0.973        & 0.021        & 0.978                             & 0.976        & 0.039        & \textbf{0.971}         \\
                & {4 clients}                                                                                          & 0.966        & 0.021        & 0.978                    & 0.971        & 0.071        & 0.957        \\
                & {10 clients}                                                                                         & 0.925        & 0.024        & \textbf{0.975}                    & 0.954        & 0.123        & 0.930         \\ \midrule
                {Centralized}                                   & {1 server}                                                                                           & 0.956        & 0.032        & 0.968                             & 0.974        & 0.038        & 0.970         \\ \bottomrule
        \end{tabular}
        }
\end{table}

\begin{figure}[!b]
  \centering
    \includegraphics[width=0.88\linewidth]{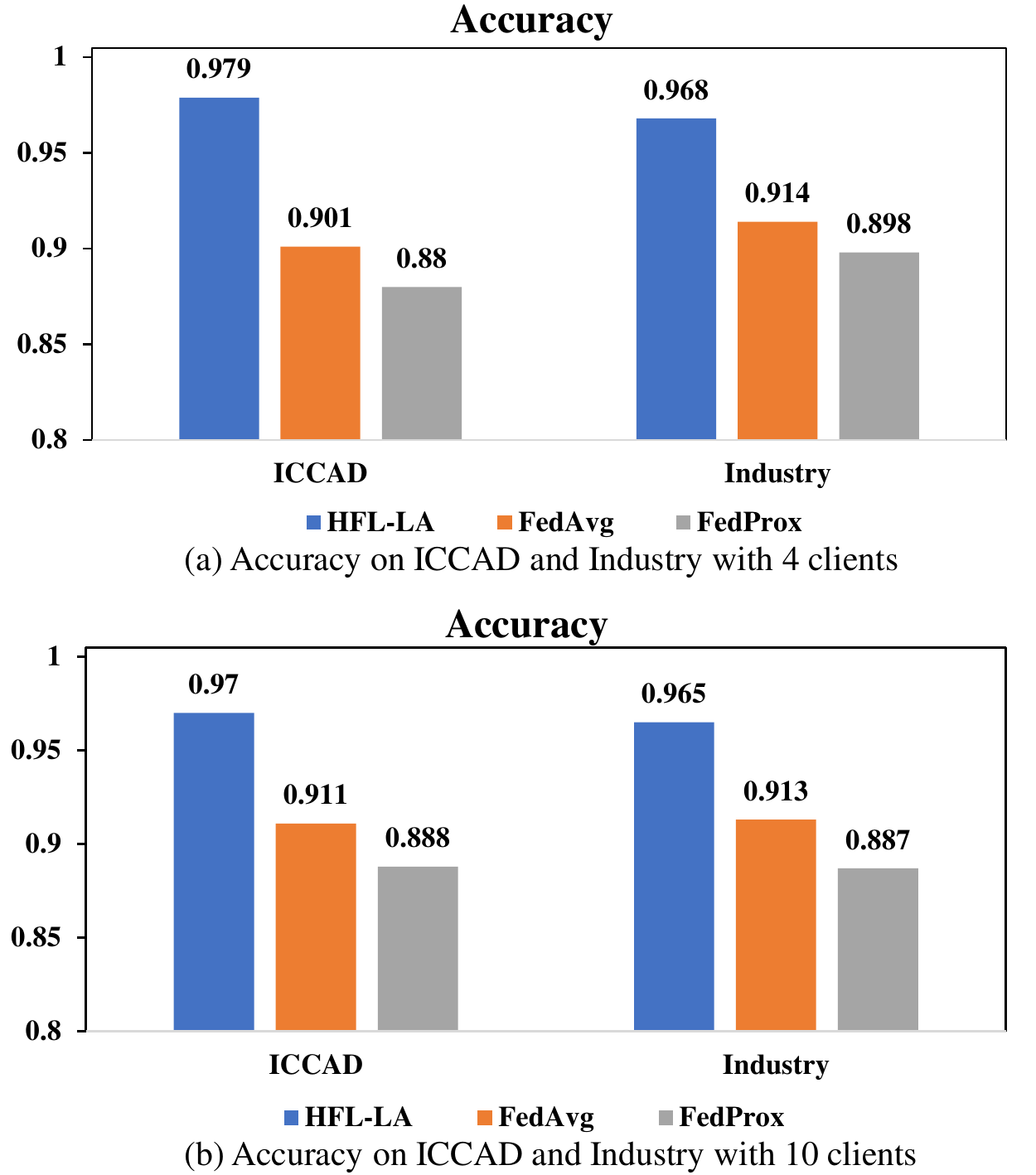}
 \caption{Accuracy comparison among HFL-LA, FedAvg, and FedProx on \texttt{ICCAD} and \texttt{Industry} with 4 and 10 clients using asynchronous model updates.}
  \label{fig:client4and10}
\end{figure}

To demonstrate the performance of the proposed HFL-LA algorithm, we compare the results of HFL-LA with that of the state-of-the-art federated learning algorithm, FedAvg in \cite{mcmahan2017communication} and FedProx in \cite{2018Federated}, as well as local and central learning. Here we have:
\begin{itemize}
    \item FedAvg: The conventional federated learning algorithm that averages over the uploaded model~\cite{mcmahan2017communication}.
    \item FedProx: The algorithm adds a proximal term to the objective to handle the heterogeneity~\cite{2018Federated}. 
    \item Local learning (denoted as local): The local learning algorithm that only uses the local data of client.
    \item Central learning (denoted as centralized): The central learning algorithm has access to all the training sets to train one unified model.
\end{itemize} In our experiments, the training sets of \texttt{ICCAD} and \texttt{Industry} benchmarks are merged together and then assigned to different numbers of clients as the local data, $i.e.$, 2, 4, and 10 clients. The testing sets have been preserved in advance as shown in Table~\ref{tab:benchmark} and used to validate the performance of the trained models. We compare the performance of the algorithms in terms of TPR, FPR, and accuracy, as defined in Sec.~III-A, and summarize the results in Table~\ref{tab:SynchronousCommunication}. In the experiments in Table~\ref{tab:SynchronousCommunication}, all the clients communicate with the server in a synchronous manner and the average of the performance across all the clients for the three scenarios of 2, 4, and 10 clients, in which the best performance cases are marked in bold. It is noted that the proposed HFL-LA can achieve 7-11\% accuracy improvement for both TPR and FPR, compared to FedAvg and FedProx. Due to the fact of using only local homogeneous training data, local learning can achieve slightly better results for \texttt{ICCAD}. However, when the data heterogeneity increases like \texttt{Industry}, the performance of local learning quickly drops and yields $\sim$4\% degradation compared to HFL-LA.

\begin{figure}[!t]
\centering
\includegraphics[width=1.\linewidth]{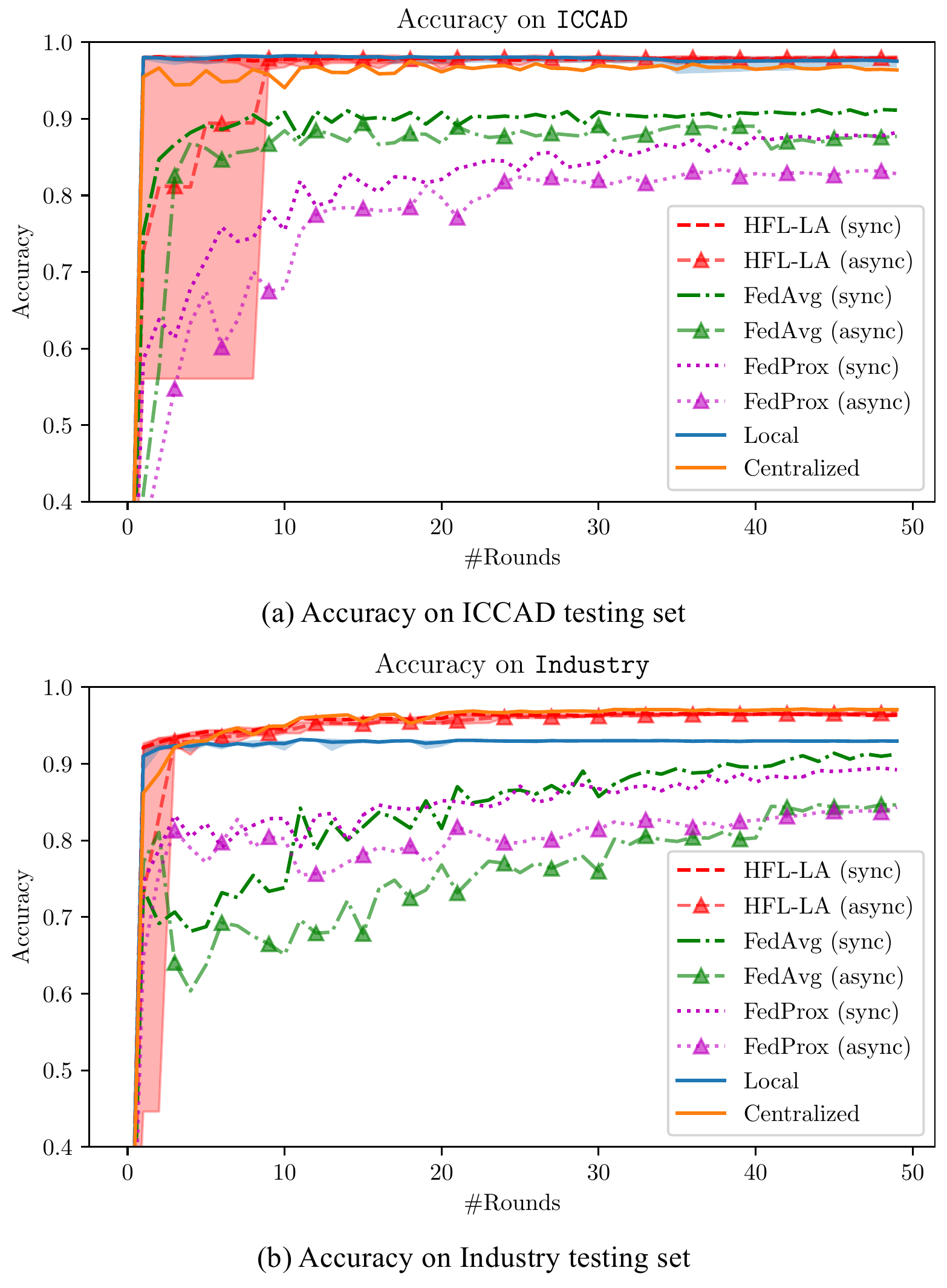}
\caption{Convergence comparison among the different methods on \texttt{ICCAD} and \texttt{Industry} during training with the models evaluated on the testing sets.}
\label{fig:acc}
\end{figure}

We further compare the results when the model can be updated asynchronously for the scenarios of 4 and 10 clients, where half  of  the  clients  are  randomly selected for training and update in each round. Since only federated learning based methods require model updates, we only compare HFL-LA with FedAvg and FedProx in Fig.~\ref{fig:client4and10}. As shown in the figure, even with heterogeneous communication and updates, HFL-LA can still achieve 5-10\% accuracy improvement from that of the other federated learning methods~\cite{mcmahan2017communication, 2018Federated}.

Finally, we compare the accuracy changes of different methods with different update mechanisms (synchronous and asynchronous, denoted as sync and async, respectively) for 10 clients during the training. 
For \texttt{ICCAD} benchmark in Fig.~\ref{fig:acc}(a), local learning and HFL-LA method achieve the highest accuracy and converge much faster than the other methods. Even with asynchronous updates, HFL-LA method can achieve convergence rate and accuracy similar to the synchronous case. For \texttt{Industry} in Fig.~\ref{fig:acc}(b), the superiority of HFL-LA is more obvious, outperforming all the other methods in terms of accuracy (e.g., 3.7\% improvement over local learning). Moreover, HFL-LA achieves almost 5$\times$ convergence speedup compared to the other federated learning methods even adopting asynchronous updates.



\section{Conclusion}

In this paper, we propose a novel heterogeneous federated learning based hotspot detection framework with local adaptation. By adopting an efficient feature selection and utilizing the domain knowledge of LHD, our framework can support the heterogeneity in data, model, and communication. Experimental results shows that our framework not only outperforms other alternative methods in terms of performance but can also guarantee good convergence even in the scenario with high heterogeneity.


\bibliographystyle{IEEEtran}

\bibliography{ref}

\begin{thebibliography}{10}
\providecommand{\url}[1]{#1}
\csname url@samestyle\endcsname
\providecommand{\newblock}{\relax}
\providecommand{\bibinfo}[2]{#2}
\providecommand{\BIBentrySTDinterwordspacing}{\spaceskip=0pt\relax}
\providecommand{\BIBentryALTinterwordstretchfactor}{4}
\providecommand{\BIBentryALTinterwordspacing}{\spaceskip=\fontdimen2\font plus
\BIBentryALTinterwordstretchfactor\fontdimen3\font minus
  \fontdimen4\font\relax}
\providecommand{\BIBforeignlanguage}[2]{{%
\expandafter\ifx\csname l@#1\endcsname\relax
\typeout{** WARNING: IEEEtran.bst: No hyphenation pattern has been}%
\typeout{** loaded for the language `#1'. Using the pattern for}%
\typeout{** the default language instead.}%
\else
\language=\csname l@#1\endcsname
\fi
#2}}
\providecommand{\BIBdecl}{\relax}
\BIBdecl

\bibitem{moore1965cramming}
G.~E. Moore \emph{et~al.}, ``Cramming more components onto integrated
  circuits,'' 1965.

\bibitem{2015Optical}
T.~Matsunawa, Y.~Bei, and D.~Z. Pan, ``Optical proximity correction with
  hierarchical bayes model,'' in \emph{Optical Microlithography XXVIII}, 2015.

\bibitem{2012Accurate}
Y.~T. Yu, Y.~C. Chan, S.~Sinha, H.~R. Jiang, and C.~Chiang, ``Accurate
  process-hotspot detection using critical design rule extraction,''
  \emph{ACM}, 2012.

\bibitem{2003Hotspot}
J.~Kim and M.~Fan, ``Hotspot detection on post-opc layout using full chip
  simulation based verification tool: A case study with aerial image
  simulation,'' \emph{Proceedings of SPIE - The International Society for
  Optical Engineering}, 2003.

\bibitem{2017Layout}
H.~Yang, S.~Jing, Z.~Yi, Y.~Ma, and E.~Young, ``Layout hotspot detection with
  feature tensor generation and deep biased learning,'' \emph{IEEE Transactions
  on Computer-Aided Design of Integrated Circuits and Systems}, vol.~PP,
  no.~99, pp. 1--1, 2017.

\bibitem{2014A}
W.~Wen, J.~Li, S.~Lin, J.~Chen, and S.~Chang, ``A fuzzy-matching model with
  grid reduction for lithography hotspot detection,'' \emph{IEEE Transactions
  on Computer-Aided Design of Integrated Circuits and Systems}, vol.~33,
  no.~11, pp. 1671--1680, 2014.

\bibitem{2015Machine}
Y.~Yu, G.~Lin, I.~H. Jiang, and C.~Chiang, ``Machine-learning-based hotspot
  detection using topological classification and critical feature extraction,''
  in \emph{IEEE}, 2015, pp. 460--470.

\bibitem{mcmahan2017communication}
B.~McMahan, E.~Moore, D.~Ramage, S.~Hampson, and B.~A. y~Arcas,
  ``Communication-efficient learning of deep networks from decentralized
  data,'' in \emph{Artificial Intelligence and Statistics}.\hskip 1em plus
  0.5em minus 0.4em\relax PMLR, 2017, pp. 1273--1282.

\bibitem{2020A}
Y.~Liu, Y.~Kang, C.~Xing, T.~Chen, and Q.~Yang, ``A secure federated transfer
  learning framework,'' \emph{Intelligent Systems, IEEE}, vol.~PP, no.~99, pp.
  1--1, 2020.

\bibitem{smith2018federated}
V.~Smith, C.-K. Chiang, M.~Sanjabi, and A.~Talwalkar, ``Federated multi-task
  learning,'' 2018.

\bibitem{2017Model}
C.~Finn, P.~Abbeel, and S.~Levine, ``Model-agnostic meta-learning for fast
  adaptation of deep networks,'' 2017.

\bibitem{2018Federated}
T.~Li, A.~K. Sahu, M.~Zaheer, M.~Sanjabi, A.~Talwalkar, and V.~Smith,
  ``Federated optimization in heterogeneous networks,'' 2018.

\bibitem{2020Think}
P.~P. Liang, T.~Liu, Z.~Liu, R.~Salakhutdinov, and L.~P. Morency, ``Think
  locally, act globally: Federated learning with local and global
  representations,'' 2020.

\bibitem{matsunawa2015optical}
T.~Matsunawa, B.~Yu, and D.~Z. Pan, ``Optical proximity correction with
  hierarchical bayes model,'' in \emph{Optical Microlithography XXVIII}, vol.
  9426.\hskip 1em plus 0.5em minus 0.4em\relax International Society for Optics
  and Photonics, 2015, p. 94260X.

\bibitem{yang2018layout}
H.~Yang, J.~Su, Y.~Zou, Y.~Ma, B.~Yu, and E.~F. Young, ``Layout hotspot
  detection with feature tensor generation and deep biased learning,''
  \emph{IEEE Transactions on Computer-Aided Design of Integrated Circuits and
  Systems}, vol.~38, no.~6, pp. 1175--1187, 2018.

\bibitem{yuan2006model}
M.~Yuan and Y.~Lin, ``Model selection and estimation in regression with grouped
  variables,'' \emph{Journal of the Royal Statistical Society: Series B
  (Statistical Methodology)}, vol.~68, no.~1, pp. 49--67, 2006.

\bibitem{2019Parallel}
H.~Yu, S.~Yang, and S.~Zhu, ``Parallel restarted sgd with faster convergence
  and less communication: Demystifying why model averaging works for deep
  learning,'' \emph{Proceedings of the AAAI Conference on Artificial
  Intelligence}, vol.~33, pp. 5693--5700, 2019.

\bibitem{paszke2019pytorch}
A.~Paszke, S.~Gross, F.~Massa, A.~Lerer, J.~Bradbury, G.~Chanan, T.~Killeen,
  Z.~Lin, N.~Gimelshein, L.~Antiga \emph{et~al.}, ``Pytorch: An imperative
  style, high-performance deep learning library,'' \emph{arXiv preprint
  arXiv:1912.01703}, 2019.

\bibitem{torres2012iccad}
J.~A. Torres, ``Iccad-2012 cad contest in fuzzy pattern matching for physical
  verification and benchmark suite,'' in \emph{2012 IEEE/ACM International
  Conference on Computer-Aided Design (ICCAD)}.\hskip 1em plus 0.5em minus
  0.4em\relax IEEE, 2012, pp. 349--350.

\end{thebibliography}

\end{document}